\documentclass[10pt,twocolumn,letterpaper]{article}

\usepackage{cvpr}
\usepackage{times}
\usepackage{epsfig}
\usepackage{graphicx}
\usepackage{amsmath}
\usepackage{amssymb}
\usepackage{subfig}
\usepackage{enumitem}
\usepackage{url}
\usepackage{textcomp}

\usepackage{array}
\newcolumntype{L}[1]{>{\raggedright\let\newline\\\arraybackslash\hspace{0pt}}m{#1}}
\newcolumntype{C}[1]{>{\centering\let\newline\\\arraybackslash\hspace{0pt}}m{#1}}
\newcolumntype{R}[1]{>{\raggedleft\let\newline\\\arraybackslash\hspace{0pt}}m{#1}}

\cvprfinalcopy 

\def\cvprPaperID{617} 

\ifcvprfinal\fi
\begin{document}

\title{Automatic Understanding of Image and Video Advertisements}


\author{Zaeem Hussain \hspace{1cm} Mingda Zhang \hspace{1cm} Xiaozhong Zhang \hspace{1cm} Keren Ye\\Christopher Thomas \hspace{1cm} Zuha Agha \hspace{1cm} Nathan Ong \hspace{1cm} Adriana Kovashka\\
Department of Computer Science\\
University of Pittsburgh\\
{\tt\small \{zaeem, mzhang, xiaozhong, yekeren, chris, zua2, nro5, kovashka\}@cs.pitt.edu }
}

\maketitle
\thispagestyle{empty}

\begin{abstract}
There is more to images than their objective physical content: for example, advertisements are created to persuade a viewer to take a certain action. We propose the novel problem of automatic advertisement understanding. To enable research on this problem, we create two datasets: an image dataset of 64,832 image ads, and a video dataset of 3,477  ads. Our data contains rich annotations encompassing the topic and sentiment of the ads, questions and answers describing what actions the viewer is prompted to take and the reasoning that the ad presents to persuade the viewer (``What should I do according to this ad, and why should I do it?''), and symbolic references ads make (e.g. a dove symbolizes peace). We also analyze the most common persuasive strategies ads use, and the capabilities that computer vision systems should have to understand these strategies. We present baseline classification results for several prediction tasks, including automatically answering questions about the messages of the ads. 
\end{abstract}



\section{Introduction}
\label{sec:intro}

\begin{figure}[t]
\includegraphics[width=1\linewidth]{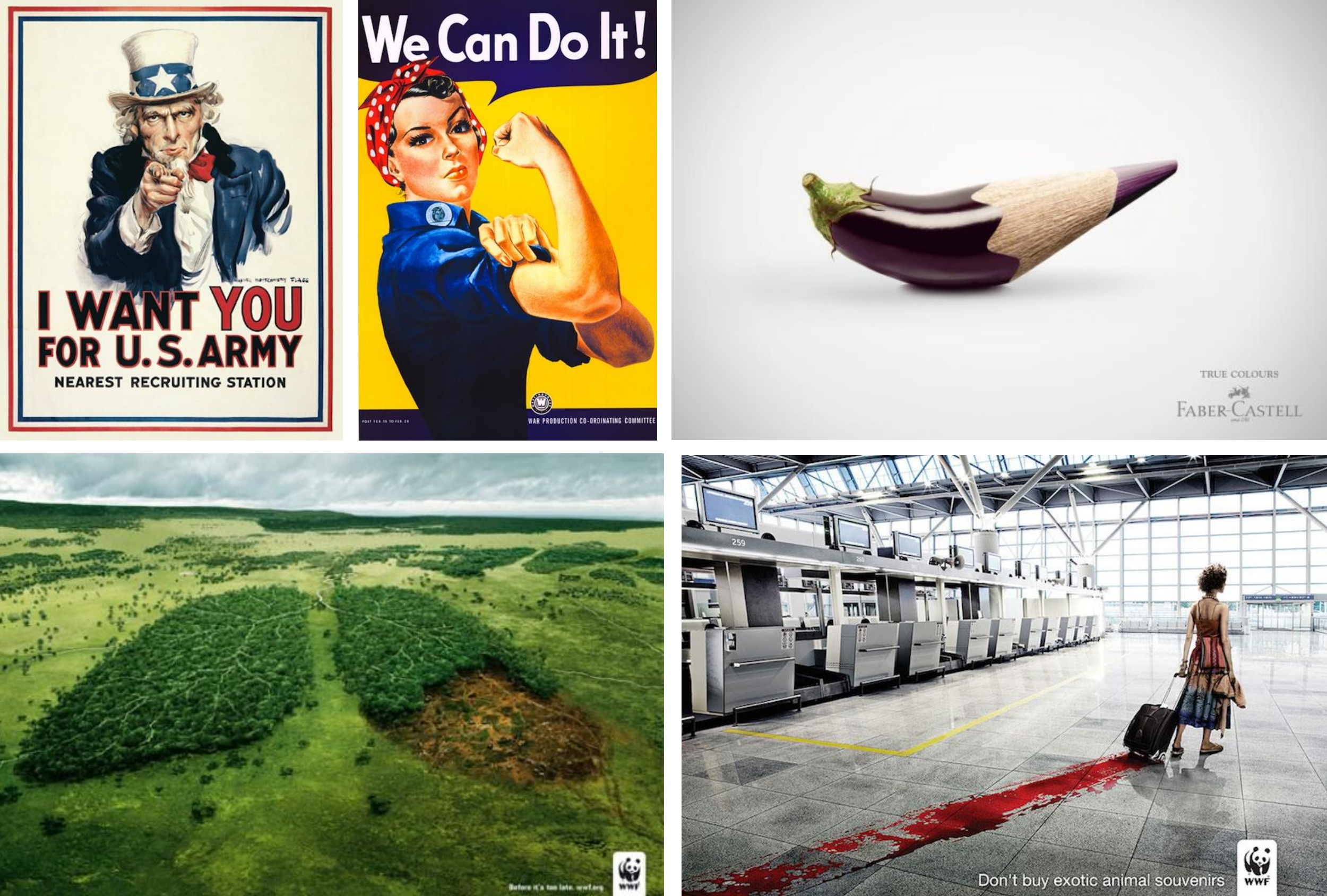}
\caption{Two iconic American ads, and three that require robust visual reasoning to decode. Despite the potential applications of ad-understanding, this problem has not been tackled in computer vision before.}
\label{fig:concept} 
\end{figure}

Image advertisements are quite powerful, and web companies monetize this power. In 2014, one fifth of Google's revenue came from their AdSense product, which serves ads automatically to targeted users \cite{adsense}. 
Further, ads are an integral part of our culture. 
For example, the two top-left ads in Fig.~\ref{fig:concept} have likely been seen by every American, and have been adapted and reused in countless ways.
In terms of video ads, 
Volkswagen's 2011 commercial ``The Force'' 
had received 8 million views before it aired on TV \cite{theforce}. 

Ads are persuasive because they convey a certain message that appeals to the viewer. 
Sometimes the message is simple, and can be inferred from body language, as in the ``We can do it'' ad in Fig.~\ref{fig:concept}.
Other ads use more complex messages, such as the inference that because the eggplant and pencil form the same object, the pencil gives a very real, natural eggplant color, as in the top-right ad in Fig.~\ref{fig:concept}. 
Decoding the message in the bottom-right ad involves even more steps, and reading the text (``Don't buy exotic animal souvenirs'') might be helpful. 
The viewer has to infer that the woman went on vacation from the fact that she is carrying a suitcase, and then surmise that she is carrying dead animals from the blood trailing behind her suitcase. 
A human knows this because she associates blood with injury or death. 
In the case of the ``forest lungs'' image at the bottom-left, lungs symbolize breathing and by extension, life. 
However, a human first has to recognize the groups of trees as lungs, which might be difficult for a computer to do.
These are just a few examples of how ads use different types of \emph{visual rhetoric} to convey their message, namely: common-sense reasoning, symbolism, and recognition of non-photorealistic objects. 
Understanding advertisements \emph{automatically} requires decoding this rhetoric. 
This is a challenging problem that goes beyond listing objects and their locations \cite{szegedy2014going, girshick2015fast,redmon2015you}, or even producing a sentence about the image \cite{vinyals2015show, donahue2015long, karpathy2015deep}, because ads are as much about \emph{how} objects are portrayed and \emph{why} they are portrayed so, as about \emph{what} objects are portrayed.

We propose the problem of ad-understanding, and develop two datasets to enable progress on it. 
We collect a dataset of over 64,000 \emph{image} ads (both product ads, such as the pencil ad, and public service announcements, such as the anti-animal-souvenirs ad). 
Our ads cover a diverse range of subjects.
We ask Amazon Mechanical Turk workers to tag each ad with its topic (e.g. what product it advertises or what the subject of the public service announcement is), what sentiment it attempts to inspire in the viewer (e.g. disturbance in the environment conservation ad), and what strategy it uses to convey its message (e.g. it requires understanding of physical processes). We also include crowdsourced answers to two questions: ``What should the viewer do according to this ad?'' and ``Why should he/she do it?''
Finally, we include any symbolism that ads use (e.g. the fact that a dove in an image might symbolically refer to the concept of ``peace''). 
We also develop a dataset of over 3,000 \emph{video} ads with similar annotations (except symbolism), and a few extra annotations (e.g. ``Is the ad funny?'' and ``Is it exciting?'') 
Our data collection and annotation procedures were informed by the literature in Media Studies, a discipline which studies the messages in the mass media, in which one of the authors has formal training.
Our data is available at 
{\tt http://www.cs.pitt.edu/\texttildelow kovashka/ads/}.
The dataset contains the ad images, video ad URLs, and annotations we collected.
We hope it will spur progress on the novel and important problem of decoding ads. 
 
In addition to creating the first pair of datasets for understanding ad rhetoric, we propose several baselines that will help judge progress on this problem. 
First, we formulate decoding ads as a question-answering problem.
If a computer vision system understood the rhetoric of an ad, it should be able to answer questions such as ``According to this ad, why should I not bully children?''
This is a very challenging task, and accuracy on it is low.
Second, we formulate and provide baselines for other tasks such as topic and sentiment recognition. 
These tasks are more approachable and have higher baseline accuracy.
Third, we 
show initial experiments on how symbolism can be used for question-answering.

The ability to automatically understand ads has many applications. 
For example, 
we can develop methods that predict how effective a certain ad will be.
Using automatic understanding of the strategies that ads use, we can help viewers become more aware of how ads are tricking them into buying certain products. 
Further, 
if we can decode the messages of ads, 
we can perform better ad-targeting according to user interests. 
Finally, decoding ads would allow us to generate descriptions of these ads for the visually impaired, and thus give them richer access to the content shown in newspapers or on TV.

\section{Related work}
\label{sec:related}

In this work, we demonstrate that there is an aspect of visual data that has not been tackled before, namely analyzing the \emph{visual rhetoric} of images. 
This problem has been studied in Media Studies
\cite{williamson1978decoding, sturken2001practices, mirzoeff2002visual, olson2008visual, o1994culture, bignell2002media, maasik2011signs, danesi2004messages}.
Further, marketing research \cite{young2005advertising} examines how viewers react to ads and whether an ad causes them to buy a product. 
While decoding ads has not been studied in computer vision, the problem is related to several areas of prior work.

\emph{Beyond objects.} 
Work on semantic visual attributes \emph{describes} images beyond \emph{labeling} the objects in them, e.g. with adjective-like properties such as ``furry'', ``smiling'', or ``metallic'' \cite{Lampert09, Farhadi09, parikh2011relative,kumar2011describable, siddiquie2011image,Kovashka15a,Kovashka15b,Fouhey16,Wang_2016_CVPR, akata2013label, jayaraman2014decorrelating}.
The community has also made first attempts in tackling content which requires subjective judgement or abstract analysis. 
For example,
\cite{pirsiavash2014assessing} learn to detect \emph{how well} a person is performing an athletic action. 
\cite{lin2015don} use the machine's ``imagination'' to answer questions about images.
\cite{karayev2013recognizing, thomas2016seeing} study the style of artistic photographs,
and \cite{doersch2012makes, lee2013style} study style in architecture and vehicles.
While these works analyze potentially subjective content, none of them analyze \emph{what the image is trying to tell us}. 
Ads constitute a new type of images, and understanding them requires new techniques.

\emph{Visual persuasion.}
Most related to our work is the visual persuasion work of \cite{joo2014visual} which analyzes whether images of politicians portray them in a positive or negative light. The authors use features that capture facial expressions, gestures, and image backgrounds to detect positive or negative portrayal. However, many ads do not show people, and even if they do, usually there is not an implication about the qualities of the person. Instead, ads use a number of other techniques, which we discuss in Sec.~\ref{sec:analysis}.

\emph{Sentiment.}
One of our tasks is predicting the sentiment an ad aims to evoke in the viewer. 
\cite{peng2015mixed, peng2016emotions, borth2013large, liu2016complura, jou2014predicting, machajdik2010affective} study the emotions shown or perceived in images, but for generic images, rather than ones purposefully created to convey an emotion. 
We compare to \cite{borth2013large} and show the success of their method does not carry over to predicting emotion in ads. 
This again shows that ads represent a new domain of images whose decoding requires novel techniques. 

\emph{Prior work on ads.}
We are not aware of any work in decoding the meaning of advertisements as we propose. 
\cite{azimi2012visual, cheng2012multimedia} predict click-through rates in ads using low-level vision features, whereas we predict what the ad is about and what message it carries. 
\cite{mcduff2014predicting} predict how much human viewers will like an ad by capturing their facial expressions. 
\cite{yadati2014cavva, mei2012imagesense} determine the best placement of a commercial in a video stream, or of image ads in a part of an image using user affect and saliency. 
\cite{sanchez2002shot, gauch2006finding} detect whether the current video shown on TV is a commercial or not, 
and \cite{sethi2013large} detect human trafficking advertisements.
\cite{zhao2011discovering} extract the object being advertised from commercials (videos), by looking for recurring patterns (e.g. logos).
Human facial reactions, ad placement and recognition, and detecting logos, are quite distinct from our goal of decoding the messages of ads.

\emph{Visual question-answering.}
One of the tasks we propose for advertisements is \emph{decoding their rhetoric}, i.e. figuring out what they are trying to say. 
We formulate this problem in the context of visual question-answering. 
The latter is a recent vision-and-language joint problem \cite{antol2015vqa, ren2015exploring, malinowski2015ask, wu2015ask, shih2015look, xu2015ask}
also related to image captioning \cite{vinyals2015show, karpathy2015deep, donahue2015long, kulkarni2013babytalk, farhadi2010every}.

\section{Image dataset}
\label{sec:image_dataset}

The first dataset we develop and make available is a large annotated dataset of image advertisements, such as the ones shown in Fig.~\ref{fig:strategies} (more examples are shown in the supplementary file).
Our dataset includes both advertisements for products, and ads that campaign for/against something, e.g. \emph{for} preserving the environment and \emph{against} bullying. 
We call the former ``product ads,'' and the latter ``public service announcements,'' or ``PSAs''.
We refer to the product or subject of the ad as its ``topic''.
We describe the image collection and annotation process below. 

\begin{figure*}[t]
\includegraphics[width=1\textwidth]{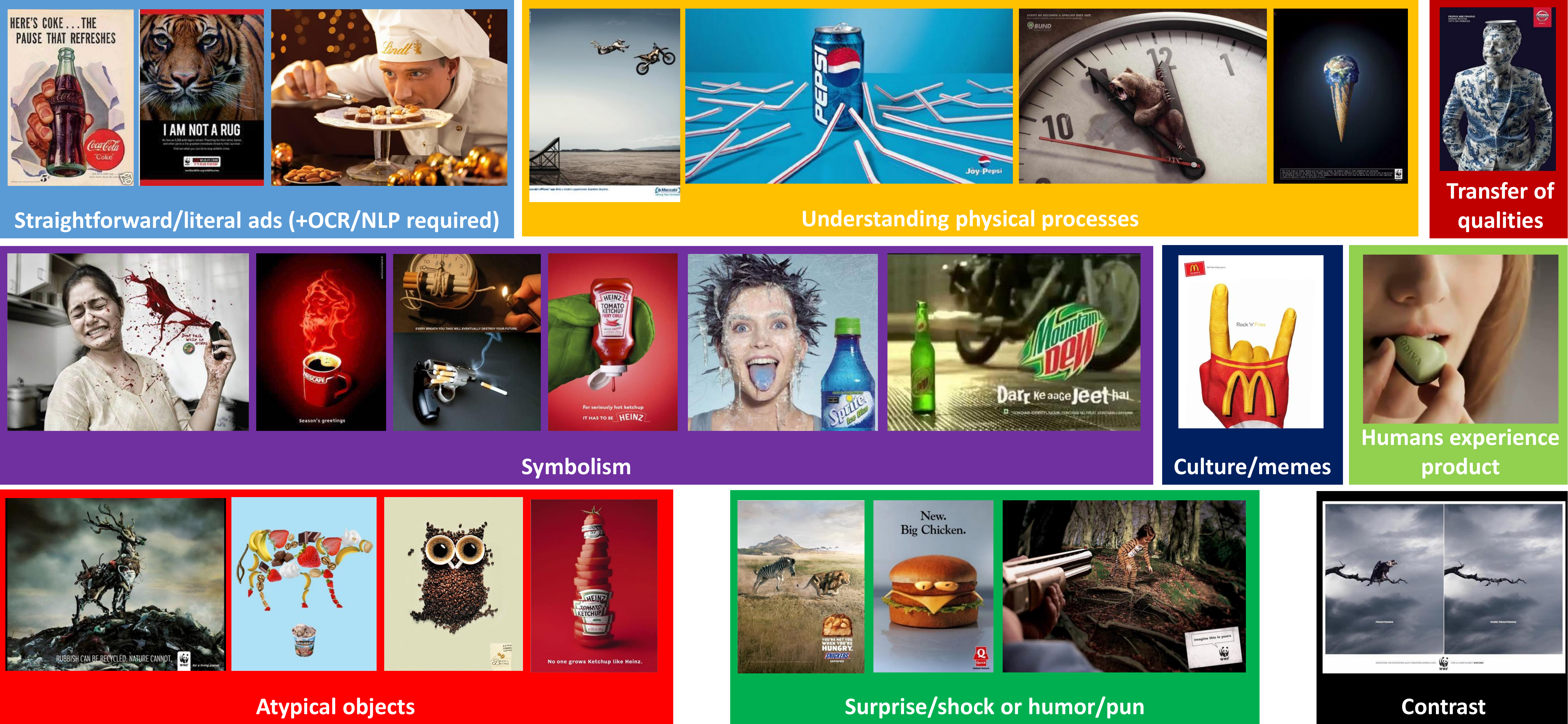}
\caption{Examples of ads grouped by strategy or visual understanding required for decoding the ad.}
\label{fig:strategies}
\end{figure*}

\subsection{Collecting ad images}

We first assembled a list of keywords (shown in supp) related to advertisements, focusing on possible ad topics. 
We developed a hierarchy of keywords that describe topics at different levels of granularity. 
This hierarchy included both coarse topics, e.g. ``fast food'', ``cosmetics'', ``electronics'', etc., as well as fine topics, such as the brand names of products (e.g. ``Sprite'', ``Maybeline'', ``Samsung''). 
Similarly, for PSAs we used keywords such as: ``smoking'', ``animal abuse'', ``bullying'', etc.
We used the entire hierarchy to query Google and retrieve all the images (usually between 600 to 800) returned for each query.
We removed all images of size less than 256x256 pixels, and obtained an initial pool of about 220,000 noisy images. 

Next, we removed duplicates from this noisy set. 
We computed a SIFT bag-of-words histogram per image, and used the chi-squared kernel to compute similarity between histograms. Any pair of images with a similarity greater than a threshold were marked as duplicates. After de-duplication, we ended up with about 190,000 noisy images. 

Finally, we removed images that are not actually advertisements, using a two-stage approach. 
First, we selected 21,945 images, and submitted those for annotation on MTurk, asking ``Is this image an advertisement? You should answer yes if you think this image could appear as an advertisement in a magazine.''
We showed plentiful examples to annotators to demonstrate what we consider to be an ``ad'' vs ``not an ad'' (examples in supp).
We marked as ads those images that at least 3/4 annotators labeled as an ad, obtaining 8,348 ads and 13,597 not-ads. 

Second, we used these to train a ResNet \cite{he2016deep} to distinguish between ads and not ads on the remaining  images. 
We set the recall of our network to 80\%, which corresponded to 85\% precision evaluated on a held-out set from the human-annotated pool of 21,945 images. 
We ran that ResNet on our 168,000 unannotated images for clean-up, obtaining about 63,000 images labeled as ads. 
We allowed annotators to label ResNet-classified ``ads'' as ``not an ad'' in a subsequent stage; annotators only used this option in 10\% of cases.
Using the automatic classification step, we saved \$1,300 in annotation costs.
In total, we obtained \textbf{64,832} cleaned-up ads. 

\subsection{Collecting image ad annotations}

We collected the annotations in Tab.~\ref{tab:image_anno}, explained below. 
Note that we describe the strategies annotations in  Sec.~\ref{sec:analysis}.

\begin{table}
\begin{tabular}{|c|c|c|}
\hline
Type & Count & Example\\
\hline
Topic & 204,340 & Electronics\\
Sentiment & 102,340 & Cheerful\\
Action/Reason & 202,090 & I should bike because it's healthy\\
Symbol & 64,131 & Danger (+ bounding box)\\
Strategy & 20,000 & Contrast\\
Slogan & 11,130 & Save the planet... save you\\
\hline
\end{tabular}
\caption{The annotations collected for our image dataset. The counts are before any majority-vote cleanup.}
\vspace{-0.3cm}
\label{tab:image_anno}
\end{table}

\vspace{-0.3cm}
\subsubsection{Topics and sentiments}

\begin{figure}
\vspace{-0.1cm}
\includegraphics[width=1\linewidth]{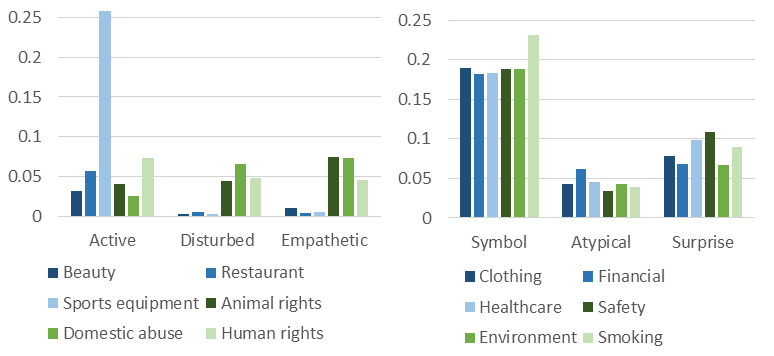}
\caption{Statistics about topics and sentiments (left), and topics and strategies (right).}
\vspace{-0.2cm}
\label{fig:stats_topic}
\end{figure}

The keyword query process used for image download does not guarantee that the images returned for each keyword actually advertise that topic. 
Thus, we developed a taxonomy of products, and asked annotators to label the images with the topic that they advertise or campaign for. 
We also wanted to know how an advertisement makes the viewer feel, since the sentiment that the ad inspires is a powerful persuasion tool \cite{mcduff2014predicting}. 
Thus, we also developed a taxonomy of sentiments.
To get both taxonomies, we first asked annotators to write free-form topics and sentiments, on a small batch of images and videos. 
This is consistent with the ``self report'' approach used to measure emotional reactions to ads \cite{poels2006capture}.
We then semi-automatically clustered them and selected a representative set of words to describe each topic and sentiment type.
We arrived at a list of 38 topics and 30 sentiments.
In later tasks, we asked workers to select a single topic and one or more sentiments. 
We collected topic annotations on all ads, and sentiments on 30,340 ads.
For each image, we collected annotations from 3 to 5 different workers.
Inter-annotator agreement on topic labels was
85\% (more details in supp).
Examples are shown in Tab.~\ref{tab:topic_sentiment}. The distribution of topics and sentiments is illustrated in Fig.~\ref{fig:stats_topic} (left); we see that sports ads and human rights ads inspire activity, while domestic abuse and human and animal rights ads inspire disturbance and empathy. Interestingly, we observe that domestic abuse ads inspire disturbance more frequently than animal rights ads do.

\begin{table}
\small
\begin{tabular}{|c|c|}
\hline
Topic  & Sentiment \\
\hline
Restaurants, cafe, fast food & Active (energetic, etc.)\\
Coffee, tea & Alarmed (concerned, etc.)\\
Sports equipment, activities & Amazed (excited, etc.) \\
Phone, TV and web providers & Angry (annoyed, irritated)\\
Education & Cheerful (delighted, etc.) \\
Beauty products & Disturbed (disgusted, shocked)\\
Cars, automobiles & Educated (enlightened, etc.) \\
Political candidates & Feminine (womanly, girlish)\\
Animal rights, animal abuse & Persuaded (impressed, etc.)\\
Smoking, alcohol abuse & Sad (depressed, etc.)\\
\hline
\end{tabular}
\caption{A sample from our list of topics and sentiments. See supp for the full list of 38 topics and 30 sentiments.}
\label{tab:topic_sentiment}
\end{table}

\vspace{-0.3cm}
\subsubsection{Questions and answers}


\begin{table}
\small
\begin{tabular}{|p{0.3\linewidth}|p{0.65\linewidth}|}
\hline
Question & Answer\\
\hline
What should you do, acc. to the ad? & I should buy Nike sportswear.\\
Why, acc. to the ad, should you do it? & Because it will give me the determination of a star athlete.\\
\hline
What? & I should buy this video game.\\
Why? & Because it is a realistic soccer experience.\\
\hline
What? & I should drink Absolut Vodka.\\
Why? & Because they support LGBT rights.\\
\hline
What? & I should look out for domestic violence.\\
Why? & Because it can hide in plain sight.\\
\hline
What? & I should not liter in the ocean.\\
Why? & Because it damages the ocean ecosystem.\\
\hline
\end{tabular}
\caption{Examples of collected question-answer pairs.} 
\label{tab:qa_examples}
\end{table}

\begin{table}
\small
\begin{tabular}{|c|c|c|c|c|c|}
\hline
\multicolumn{3}{|c|}{\emph{What} should you do?} & \multicolumn{3}{|c|}{\emph{Why} should you do it?}\\
\hline
Educat. & Travel & Smoking & Educat. & Travel & Smoking\\
\hline
go & go & smoke & help & fun & smoking\\
college & visit & cigarette & learn & beautiful & like\\
use & fly & buy & want & like & kill\\
attend & travel & stop & career & want & make\\
school & airline & quit & things & great & life\\
\hline
\end{tabular}
\caption{Common words in responses to action and reason questions for selected topics, from the image dataset.} 
\label{tab:qa_words_examples}
\end{table}

We collected 202,090 questions and corresponding answers, with three question-answer pairs per image. Tab.~\ref{tab:qa_examples} shows a few examples. 
We asked MTurk workers ``What should you do, according to this ad, and why?''
The answer then describes the message of the ad, e.g. ``I should buy this dress because it will make me attractive.''
We required workers to provide answers in the form ``I should [Action] because [Reason].''
Since the question is always the same, we automatically reformatted the annotator's answer into a question-answer pair, as follows. 
The question became ``Why should you [Action]?'' and the answer became ``Because [Reason].'' 
For later tasks, we split this into \emph{two} questions, i.e. we separately asked about the ``What?'' and the ``Why?'' 
However, Tab.~\ref{tab:image_anno} counts these as a single annotation.
Examples of the most commonly used words in the questions and answers are shown in Tab.~\ref{tab:qa_words_examples}.

\vspace{-0.3cm}
\subsubsection{Symbols}

In the second row of Fig.~\ref{fig:strategies}, the first image uses blood to symbolize injury, the second symbolically refers to the holiday spirit via the steam, the third uses a gun to symbolize danger, the fourth uses an oven mitt to symbolize hotness, the fifth uses icicles to symbolize freshness, and the sixth uses a motorbike to symbolize adventure.
Decoding symbolic references is difficult because it relies on human associations.
In the Media Studies literature, the physical object or content that stands for some conceptual symbol is called ``signifier'', and the symbol is the ``signified'' \cite{williamson1978decoding}.

We develop a list of symbols  (concepts, signifieds) and corresponding training data, using the help of MTurkers. 
We use a two-stage process. 
First, we ask annotators whether an ad can be interpreted literally (i.e. is straight-forward), or it requires some non-literal  interpretation. 
For simplicity, we treat all non-literal strategies as symbolism.
If the majority of MTurkers respond the ad is non-literal, it enters a second stage, in which we ask them to label the signifier and signified. 
In particular, we ask them to draw a bounding box (which denotes the signifier)
and label it with the symbol it refers to (the signified).
13,938 of all images were found to contain symbolism. 
We prune extremely rare symbols and arrive at a list of 221 symbols, each with a set of bounding boxes. 
The most common symbols are: ``danger,'' ``fun,'' ``nature,'' ``beauty,'' ``death,'' ``sex,'' ``health,'' and ``adventure.''
More statistics are in supp.

\vspace{-0.3cm}
\subsubsection{Slogans}

Additionally, for a small number of ads, we also asked MTurkers to write creative slogans that capture the message of the ad.
While we did not use this data in our work, we obtained some intriguing answers, which we think can inspire interesting work on slogan generation.


\subsection{Challenges of collection and quality control}

A data collection task of this magnitude presented challenges on three fronts: speed of collection, cost, and quality. 
For each kind of annotation, we started with a price based on the estimated time it took to complete the task. As results would come in, we would adjust this price to account for the actual time taken on average and, often, also to increase the speed with which the tasks were being completed. Even after increasing the pay significantly, some of the more difficult tasks, such as identifying symbolism and question-answering, would still take a long time to complete. For symbolism, we offered a bonus to MTurkers who would do a large number of tasks in one day. 
In total, collecting annotations for both image and video ads cost \textbf{\$13,030}. 

For the tasks where MTurkers just had to select options, such as topics and sentiments, we relied on a majority vote to disregard low-quality work. For question-answering, we used heuristics, the number of short or repetitive responses and number of non-dictionary words in the answers, to shortlist suspicious responses. For symbolism, we manually reviewed a random subset of responses from each MTurker who did more than a prespecified number of tasks in a day.

\section{How can we decode ads?}
\label{sec:analysis}

What capabilities should our computer vision systems have, in order to automatically understand the messages that ads convey, and their persuasive techniques? 
For example, would understanding the ad be straightforward if we had perfect object recognition?
We develop a taxonomy that captures the key strategies that ads use.
While ad strategy and type of visual understanding are not the same, they influence each other, so our analysis captures both.

Five of the authors each labeled 100 ads with the strategy the ad uses. 
We did so using a shared spreadsheet where we progressively added new strategies in free-form text, as we encountered them, or selected a previously listed strategy. 
After all 5x100 images were annotated, one author checked for consistency and iteratively merged similar strategies, resulting in a final list of nine strategies shown in Fig.~\ref{fig:strategies}:
\begin{itemize}[itemsep=3pt,topsep=3pt,parsep=0pt,partopsep=0pt]
\item \emph{Straightforward/literal} ads that only require object recognition and text recognition and understanding;
\item Ads that imply some dynamic \emph{physical process} is taking place, and this process is the reason why the product is valuable (e.g. the straws are striving towards the can) or why action must be taken (e.g. the arms of the clock are crushing the bear, so time is running out); 
\item Ads where \emph{qualities} of one object (e.g. the fragility of a teacup) \emph{transfer} to another (the person);
\item Ads where an object \emph{symbolizes} an external concept;
\item Ads that make references to \emph{cultural} knowledge;
\item Ads that illustrate the qualities of a product with a person \emph{experiencing} them;
\item Ads that show \emph{atypical} non-photorealistic objects;
\item Ads that convey their message by \emph{surprising, shocking} or entertaining the viewer through \emph{humor};
\item Ads that demonstrate the qualities of products, or dangers of environmental processes, through \emph{contrast}.
\end{itemize}

\begin{figure}
\centering
\includegraphics[width=0.8\linewidth]{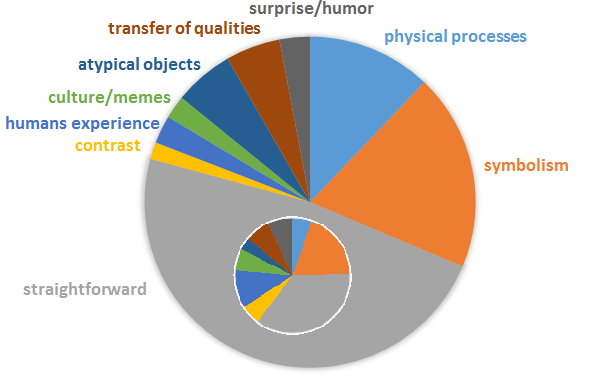}
\caption{Strategies and visual understanding statistics. The main figure shows annotations from the authors; the inset shows annotations from MTurk workers. Best in color.}
\label{fig:strategies_chart}
\end{figure}

Each image could be labeled with multiple strategies. 
We computed what fraction of the total number of strategy instances across ads belong to each strategy.
We illustrate the result in the main chart in Fig.~\ref{fig:strategies_chart}. 
In order to compute statistics over more ads, we also asked MTurk workers to label the strategy for a set of 4000 ads. 
We obtained a similar chart (shown as the inset in Fig.~\ref{fig:strategies_chart}) where the more rare strategies appeared slightly more commonly, likely because different viewers have a different internal ``clustering'' of strategies, and our MTurk annotators are not vision experts. 
In both the authors' pie chart and the crowdsourced one, straightforward ads and symbolic ads are most common.

Based on the statistics in Fig.~\ref{fig:strategies_chart}, the straightforward strategy which can be decoded with perfect object recognition accounts for less than 50\% of all strategy instances. Thus, as a community, we also must tackle a number of other challenges summarized below, to enable ad-decoding. 
Note that decoding ads involves a somewhat unique problem: the number of ads for each strategy is not very large, so applying standard deep learning techniques may be infeasible.

\begin{itemize}[itemsep=3pt,topsep=3pt,parsep=0pt,partopsep=0pt]
    \item We need to develop a method to decode symbolism in ads. We make an initial attempt at this task, in Sec.~\ref{sec:symbolism_expts}.
    \item We need to develop techniques to understand physical processes in ads (e.g. the straws \emph{striving towards} the can, or the bear \emph{being crushed}). There is initial work in understanding physical forces \cite{mottaghi2016happens, mottaghi2016newtonian, zheng2014detecting} and object transformations  \cite{isola2015discovering}, but this work is still in its infancy and is not sufficient for full physical process understanding, as we need for many ads.
    \item We need robust algorithms that can recognize objects in highly non-photorealistic modalities. For example, vision algorithms we experimented with were unable to recognize the deer, cow, owl and bottle under ``Atypical objects'' in Fig.~\ref{fig:strategies}. This may be because here these objects appear with very distinct texture from that seen in training images. There is work in learning domain-invariant representations \cite{gong2012geodesic,tzeng2015simultaneous, bousmalis2016domain,castrejon2016learning,ganin2015unsupervised,ghifary2016deep,kan2015bi,long2016unsupervised,christoudias2010learning}, but a challenge in the case of ads is that data from each ``domain'' (the particular way e.g. the deer is portrayed) may be limited to single examples.  
    \item We need techniques to understand what is surprising or funny in an ad. There is initial work on abnormality  \cite{saleh2016toward,wang2016what} restricted to modeling co-occurrences of objects and attributes, and on humor  \cite{chandrasekaran2016we} in cartoons, but surprise/humor detection remains largely unsolved.
\end{itemize}

Finally, we also analyze correlations between ad topics and ad strategies, in Fig.~\ref{fig:stats_topic} (right). We see that symbols are used in a variety of ads, but most commonly in smoking ads. Financial ads use atypical portrayals of objects most frequently, and healthcare and safety ads use surprise. 
\section{Video dataset}
\label{sec:video_dataset}

Video advertisements are sometimes even more entertaining and popular than image ads. 
For example, an Old Spice commercial\footnote{\url{https://www.youtube.com/watch?v=owGykVbfgUE}} has over 54 million views. 
However, commercials are expensive to make, and might cost several million USD to air \cite{fortune}. 
Thus, there are fewer commercials available on the web, hence our video dataset is smaller. 

\subsection{Collecting ad videos}

We obtained a list of 949 videos from an Internet service provider. 
However, we wanted to increase the size of the dataset, so we additionally crawled YouTube for videos, using the keywords we used to crawl Google for images. We picked videos that have been played at least 200,000 times and have more ``likes'' than ``dislikes''.
We ran an automatic de-duplication step.
For every video, we separately took (1) 30 frames from the beginning and (2) 30 from the end, lowered their resolution, then averaged over them to obtain a single image representation, which is presumably less sensitive to slight variations. If both the start and end frames of two videos matched according to a hashing algorithm \cite{zauner2010implementation}, they were declared duplicates. 
We thus obtained an additional set of 5,028 noisy videos, of which we submitted 3,000 for annotation on Mechanical Turk. 
We combined the ad/not ad cleanup with the remainder of the annotation process.
We used intuitive metrics to ensure quality, e.g. we removed videos that were low-resolution, very old, spoofs, or simply not ads. We thus obtained \textbf{3,477} video ads in total.

\subsection{Collecting video ad annotations}

\begin{figure}
\includegraphics[width=1\linewidth]{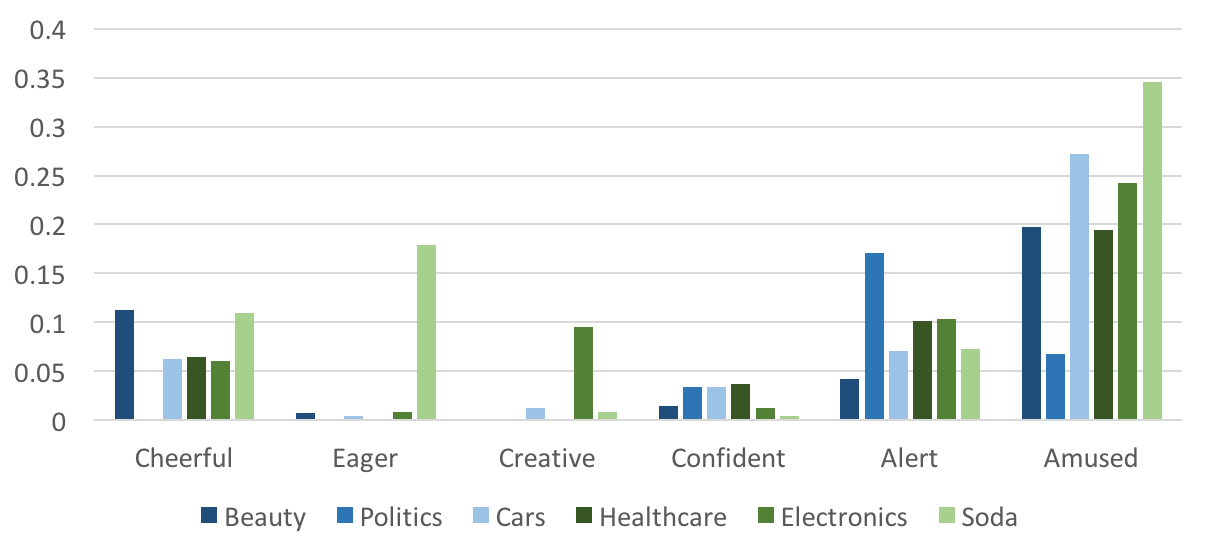}
\caption{Statistics of the video dataset.}
\label{fig:stats_video}
\end{figure}

We collected the types of annotations shown in Tab.~\ref{tab:video_anno}. 
We showed workers examples for how to annotate, on six videos.  
The topic and sentiment multiple-choice options overlap with those used for images. 
We also obtained answers to the questions ``What should you do according to this video?'' and ``Why should you do this, according to the video?''
We show statistics in Fig.~\ref{fig:stats_video} and Tab.~\ref{tab:qa_video_words_examples}, and more in supp.
For example, we see cheerfulness is most common for beauty and soda ads, eagerness for soda ads, creativeness for electronics ads, and alertness for political ads.

Our video dataset has two additional annotations, namely whether a video ad is ``funny'' or ``exciting''. Since a video has more space/time to convey its message, we thought humor and excitement are more typical for video ads. 
In contrast, we found symbolism less typical.


\begin{table}
\begin{tabular}{|c|c|C{4cm}|}
\hline
Type & Count & Example\\
\hline
Topic & 17,345 & Cars/automobiles, Safety \\
Sentiment & 17,345  & Cheerful, Amazed \\
Action/Reason & 17,345 & I should buy this car because it is pet-friendly\\
Funny? & 17,374 & Yes/No \\
Exciting? & 17,374 & Yes/No  \\
English? & 15,380 & Yes/No/Does not matter \\
Effective? & 16,721 & Not/.../Extremely Effective \\
\hline
\end{tabular}
\caption{The annotations collected for our video ad dataset.}
\label{tab:video_anno}
\end{table}

\begin{table}
\small
\begin{tabular}{|c|c|c|c|c|c|}
\hline
\multicolumn{3}{|c|}{\emph{What} should you do?} & \multicolumn{3}{c|}{\emph{Why} should you do it?}\\
\hline
Educat. & Travel & Charity & Educat. & Travel & Charity\\
\hline
univ. & visit & support & help & fun & help\\
enroll & go & donate & get & family & cancer\\
college & vacation & charity & degree & travel & need\\
online & travel & money & univ. & place & children\\
attend & use & foundat. & offer & vacation & people\\
\hline
\end{tabular}
\caption{Common responses to action and reason questions.}
\label{tab:qa_video_words_examples}
\end{table}

\section{Experiments}
\label{sec:results}

Thus far, we described our collected data, and analysis on what is required in order to be able to decode ads. 
We now describe our evaluation of baselines for several prediction tasks on ads. 
For most prediction tasks we treat the possible labels as mutually exclusive, and report accuracy. For symbolism detection, we predict multiple labels per image, and report the overall F-score.

\subsection{Question-answering for image ads}
\label{sec:image_qa_results}

We first evaluate how well an existing question-answering method performs on our questions about ads. 
In order to answer questions about ads, computer vision systems need to understand the implicit visual rhetoric and persuasion techniques of ads. 
We show existing methods do \emph{not} have this capability, via their low performance. 

Because our amount of data is limited (64,832 images with 3 or 5 question-answer pairs per image), we opted for a simple question-answering approach \cite{antol2015vqa}, trained on our ads data. We use a two-layer LSTM to encode questions in 2048D, and the last hidden layer of VGGNet \cite{simonyan2014very} to encode images in 4096D. We then project these to 1024D each, concatenate them, and add a 1000D softmax layer, which is used to generate a single-word answer. For each image, we have three reformatted questions about the persuasive strategy that the ad uses, of the type ``Why should you [Action]?'' where Action can be e.g. ``buy a dress''. 
We select one of the three questions for training/testing, namely the one whose words have the highest average TFIDF score. 
The TFIDF scores are calculated based on all
questions and answers.
We pair that question with modified versions of the three answers that annotators provided, of the form ``Because [Reason],'' where Reason can be e.g. ``it will make me pretty.'' 
Since our data originally contains \emph{sentence} (not single-word) answers, we trim each of the three answers to the single most ``contentful'' word, i.e. the word that has the highest TFIDF score.

We consider the predicted answer to be correct if it matches any of the three human answers.
Frequently our human annotators provide different answers, due to synonymy or because they interpreted the ad differently. 
This is in contrast to more objective QA tasks \cite{antol2015vqa} where answers are more likely to converge. 
Similarly, the QA method might predict a word that is related to the annotator answers but not an exact match. 
Thus, our QA task is quite challenging. 
Using the approach described, we obtain \textbf{11.48\%} accuracy.
This is lower than the accuracy of the ``why'' questions from the original VQA
(using a different test setup). 


To simplify the QA prediction task, we also conducted an experiment where we clustered the original full-sentence answers into 30 ``prototype'' answers, and trained a network to predict one of the 30 cluster IDs. The intuition is that while there are many different words annotators use to answer our ``Why'' questions, there are common patterns in the reasons provided. The baseline network (using a 128D question encoding and 512D image encoding) achieved \textbf{48.45\%} accuracy on this task. 
We next attempt to improve these numbers via symbolism decoding.

\subsection{Symbolism prediction}
\label{sec:symbolism_expts}

We use an attention model \cite{sharma2012discriminative, jetley2016end} to detect symbols; we found that to work slightly better than direct classification. Details of the model are provided in supp.  
Multiple symbols might be associated with different regions in the image.
The network achieves F-score of \textbf{15.79\%} in distinguishing between the 221 symbols. 

Note that learning a symbol detector is very challenging due to the variable data within each symbol. For example, the symbolic concept of ``manliness'' can be illustrated with an attractive man posing with a muscle car, a man
surrounded by women, etc.
We show some examples in supp.

To improve symbol predictions, we also experimented with grouping symbols into clusters based on synonymy and co-occurrence, obtaining 53 symbols (see supp for details). 
A model trained to distinguish between these 53 symbols achieved \textbf{26.84\%} F-score.

We also did a preliminary experiment using symbolism  for question-answering. 
For the 1000-way single-word prediction task, we used the class probability of each symbol as an extra feature to our QA network, and obtained slightly improved accuracy of \textbf{11.96\%} (compared to 11.48\% for the baseline).
On the 30-way QA task, a method which replaced the baseline's image features with 3x512D ones obtained from a network fine-tuned to predict (1) symbols, (2) topics, and (3) sentiments, achieved \textbf{50\%} accuracy (compared to 48.45\%). 
Devising a better way to predict and use symbolism for question-answering is our future work.

\subsection{Question-answering for video ads}
\label{sec:video_qa_results}

We used the same process as above and the video features from Sec.~\ref{sec:video_topic_results}. We achieved QA accuracy of \textbf{8.83\%}. 

\subsection{Topic and sentiment on image ads}
\label{sec:image_topic_results}

We chose the most frequent topic/sentiment as the ground-truth label.
We trained  152-layer ResNets \cite{he2016deep} to discriminate between our 38 topics and 30 sentiments. The network trained on topics achieved \textbf{60.34\%} accuracy on a held-out set.
The sentiment network achieved \textbf{27.92\%} accuracy. 
Thus, predicting the topic of an ad is much more feasible with existing techniques, compared to predicting the message as in the QA experiments above.

For sentiments, we also trained a classifier on the data from the Visual Sentiment Ontology \cite{borth2013large}, to establish how sentiment recognition on our ads differs from recognizing sentiments on general images. We map \cite{borth2013large}'s Adjective Noun Phrases (ANPs) to the sentiments in our data, by retrieving the top 10 ANPs closest to each of our sentiment words, measuring cosine similarity in word2vec space \cite{mikolov2013distributed, mikolov2013linguistic}.
We use images associated with all ANPs mapped to one of our sentiments, resulting in 21,523 training images (similar to our number of sentiment-annotated images). 
This achieves 
\textbf{6.64\%} accuracy, 
lower than the sentiment accuracy on our ad data, indicating that sentiment on ads looks different than sentiment on other images.


\subsection{Topic and sentiment on video ads}
\label{sec:video_topic_results}

We believe the actions in the video ads may have significant impact on understanding the ads.
Thus, we used the C3D network \cite{tran2015learning,  karpathy2014large} originally used for action recognition as a feature extractor.
It is pre-trained on Sports-1M \cite{karpathy2014large} and fine-tuned on UCF101 \cite{soomro2012ucf101}.
We converted videos into frames, and took consecutive 16 frames as a clip. 
We extracted fc6 and fc7 features for each clip and simply averaged the features for all clips within the same video.

We trained separate multi-class SVMs to distinguish between our 38 topics and 30 sentiments. We found fc7 shows better performance. With the optimal parameters from a validation set, we achieved \textbf{35.1\%} accuracy for predicting video topics, and \textbf{32.8\%} accuracy for sentiments. We had limited success with directly training a network for this task. 


\subsection{Funny/exciting for video ads}
\label{sec:video_funny_results}

We used a similar strategy to predict ``funny'' and ``exciting'' binary labels on videos. We excluded videos that are ambiguous (i.e. obtained split positive/negative votes).
We trained binary SVMs on the fc7 features, and obtained  \textbf{78.6\%} accuracy for predicting humor, and \textbf{78.2\%} for predicting excitement. Note that a majority class baseline achieves only 58\% and 60.8\%, respectively. Thus, predicting humor and excitement is surprisingly feasible. 


\section{Conclusion}
\label{sec:conclusion}

We have proposed a large annotated image advertisement dataset, as well as a companion annotated video ads dataset. We showed analysis describing what capabilities we need to build for vision systems so they can understand ads, and showed an initial solution for decoding symbolism in ads. 
We also showed baselines on several tasks, including question-answering capturing the subtle messages of ads.

We will pursue several opportunities for future work. We will further develop our symbolism detection framework, including additional weakly labeled web data for each symbol. We will also make use of knowledge bases for decoding ads. 
We will model video advertisements with an LSTM network and better features, and include audio processing in our analysis. 
We will use the topic, sentiment, humor and excitement predictions to improve the accuracy of question-answering. 
Finally, we will also pursue recognizing atypical objects and modeling physical processes.

\vspace{-0.3cm}
\paragraph{Acknowledgements:} 
This material is based upon work supported by the National Science Foundation under Grant Number 1566270. 
This research was also supported by a Google Faculty Research Award and an NVIDIA hardware grant. 
Any opinions, findings, and conclusions or recommendations expressed in this material are those of the author(s) and do not necessarily reflect the views of the National Science Foundation.

{\small
\bibliographystyle{ieee}
\bibliography{refs}
}

\end{document}